\documentclass[11pt,a4paper]{article}
\usepackage[hyperref]{emnlp-ijcnlp-2019}
\usepackage{times}
\usepackage{latexsym}
\usepackage{url}
\usepackage{graphicx}
\usepackage{booktabs}
\usepackage{xcolor}
\usepackage{comment}
\usepackage{amsmath}
\usepackage{amsfonts}
\usepackage{upgreek}
\usepackage{siunitx}
\usepackage{multirow}

\usepackage{natbib}
\usepackage{xspace}
\usepackage{soul}
\usepackage{adjustbox}
\usepackage{wrapfig}

\aclfinalcopy 

\title{Interactive Language Learning by Question Answering}

\author{Xingdi Yuan$^\heartsuit$\thanks{\:\:\:\:Equal contribution.} \:\:\:\: Marc-Alexandre C\^ot\'{e}$^\heartsuit$\footnotemark[1] \:\:\:\: Jie Fu$^{\clubsuit\spadesuit}$ \:\:\:\: Zhouhan Lin$^{\diamondsuit\spadesuit}$\\ \textbf{Christopher Pal$^{\clubsuit\spadesuit}$ \:\:\:\: Yoshua Bengio$^{\diamondsuit\spadesuit}$ \:\:\:\: Adam Trischler$^\heartsuit$}\\
$^\heartsuit$Microsoft Research, Montr\'{e}al \:\:\:\: $^\clubsuit$Polytechnique Montr\'{e}al \:\:\:\: $^\diamondsuit$Universit\'{e} de Montr\'{e}al \:\:\:\:$^\spadesuit$ Mila\\
eric.yuan@microsoft.com\:\:\:\:macote@microsoft.com
}

\date{}

\newcounter{lintr}

\newcommand{\smx}{{\mathrm{softmax}}}

\newcommand{\mdqn}{QA-DQN\xspace}

\newcolumntype{R}[2]{%
    >{\adjustbox{angle=#1,lap=\width-(#2)}\bgroup}%
    l%
    <{\egroup}%
}
\newcommand*\rot{\multicolumn{1}{R{70}{1em}}}

\definecolor{color1}{HTML}{da6752}
\definecolor{color2}{HTML}{5573a6}
\definecolor{color3}{HTML}{6f9f6a}
\definecolor{color4}{HTML}{f3905c}

\newcommand{\qait}{QAit\xspace}

\newcommand{\code}[1]{\texttt{#1}}
\newcommand{\cmd}[1]{\textbf{\small{\code{#1}}}}
\newcommand{\cmds}[1]{\textbf{\scriptsize{\code{#1}}}}

\begin{document}
\maketitle
\begin{abstract}
Humans observe and interact with the world to acquire knowledge.
However, most existing machine reading comprehension (MRC) tasks miss the interactive, information-seeking component of comprehension. Such tasks present models with static documents that contain all necessary information, usually concentrated in a single short substring. Thus, models can achieve strong performance through simple word- and phrase-based pattern matching.
We address this problem by formulating a novel text-based question answering task: Question Answering with Interactive Text (\qait).\footnote{The dataset and implementation of our baseline agents are publicly available at \url{https://github.com/xingdi-eric-yuan/qait_public}.}
In \qait, an agent must interact with a partially observable text-based environment to gather information required to answer questions.
\qait poses questions about the existence, location, and attributes of objects found in the environment.
The data is built using a text-based game generator that defines the underlying dynamics of interaction with the environment.
We propose and evaluate a set of baseline models for the \qait task that includes deep reinforcement learning agents.
Experiments show that the task presents a major challenge for machine reading systems, while humans solve it with relative ease.
\end{abstract}

\section{Introduction}
\label{section:intro}

The research community has defined the task of machine reading comprehension (MRC) to teach machines to read and understand text. In most MRC tasks, given a knowledge source (usually a text document) and a question on its content, a model is required to answer the question either by pointing to words in the source or by generating a text string.
Recent years have seen a flourishing of MRC works, including the release of numerous datasets \cite{rajpurkar16squad, trischler16newsqa, nguyen16msmarco, reddy18coqa, yang18hotpot} and a host of neural models that address them \cite{wang16matchlstm, wang17rnet, chen17drqa}.

\begin{table}[t!]
    \scriptsize
    \centering
    \begin{tabular}{r|l}
        \toprule
        Game        &  You find yourself in a backyard.  You make out a patio table. \\
                    &  You see a patio chair. It is stylish. In a corner you can \\
                    &  see a BBQ. There is an open screen door leading south. \\
        \midrule
        Question    &   Is ghargh edible? \\
        \midrule
        \midrule 
        Agent       &   \cmds{$>$ Qapla'} \\
        \midrule
        Game        &   That's not a verb I recognize.\\
        \midrule
        Agent       &   \cmds{$>$ eat ghargh} \\
        \midrule
        Game        &   You can't see any such thing. \\
        \midrule
        Agent       &   \cmds{$>$ go south} \\
        \midrule
        Game        &  You've just entered a kitchen. You can see an open fridge \\
                    &  here. It is empty! You see a closed oven. Look over there, it's \\
                    &  a counter. On the counter you can see a diced potato, a red\\
                    &  ghargh and a knife. There is an open screen door leading north. \\
        \midrule
        Agent       &   \cmds{$>$ eat ghargh} \\
        \midrule
        Game        &   You need to take the red ghargh first.\\
        \midrule
        Agent       &   \cmds{$>$ take red ghargh} \\
        \midrule
        Game        &   You take the red ghargh from the counter.\\
        \midrule
        Agent       &   \cmds{$>$ eat ghargh} \\
        \midrule
        Game        &   You eat the red ghargh. Not bad.\\
        \midrule
        Agent       &   \cmds{$>$ wait}\\
        \midrule
        Agent       &   \cmds{$>$ answer: yes}\\
        \bottomrule
    \end{tabular}
    \caption{Example of a player interacting with an environment to gather knowledge and answer a question.}
    \label{tab:interactiveqa_example}
\end{table}


Most existing MRC datasets, as a consequence of their construction, encourage models to do fairly shallow phrase and word matching between the question and the knowledge source.
There are three reasons for this.
First, questions tend to paraphrase or permute the text that expresses the answer, making pattern matching an effective strategy.
On the SQuAD development set, for example, the average precision, recall, and $F_1$ score between questions and the sentences containing answer spans are 0.409, 0.271, 0.301, respectively. 
Exploiting this overlap, \citet{trischler16newsqa} show that a simple Information Retrieval method can achieve high sentence-level accuracy on SQuAD.

Second, the information that supports predicting the answer from the source is often \emph{fully observed}:
the source is static, sufficient, and presented in its entirety.
This does not match the information-seeking procedure that arises in answering many natural questions~\cite{kwiatkowski19naturalquestions}, nor can it model the way humans observe and interact with the world to acquire knowledge.

Third, most existing MRC studies focus on \emph{declarative} knowledge --- the knowledge of facts or events that can be stated explicitly (i.e., declared) in short text snippets.
Given a static description of an entity, declarative knowledge can often be extracted straightforwardly through pattern matching. For example, given the EMNLP website text, the conference deadline can be extracted by matching against a date mention.
This focus overlooks another essential category of knowledge --- \emph{procedural} knowledge.
Procedural knowledge entails executable sequences of actions. These might comprise the procedure for tying ones shoes, cooking a meal, or gathering new declarative knowledge. The latter will be our focus in this work. As an example, a more general way to determine EMNLP's deadline is to open a browser, head to the website, and then match against the deadline mention; this involves executing several mouse and keyboard interactions.
In order to teach MRC systems \emph{procedures} for question answering, we propose a novel task: Question Answering with Interactive Text (\qait).
Given a question $q \in Q$, rather than presenting a model with a static document $d \in D$ to read, \qait requires the model to interact with a partially observable environment $e \in E$ over a sequence of turns. The model must collect and aggregate evidence as it interacts, then produce an answer $a$ to $q$ based on its experience.


In our case, the environment $e$ is a text-based game with no explicit objective. The game places an agent in a simple modern house populated by various everyday objects. The agent may explore and manipulate the environment by issuing text commands.
An example is shown in Table~\ref{tab:interactiveqa_example}.
We build a corpus of related text-based games using a generator from~\citet{cote18textworld}, which enables us to draw games from a controlled distribution.
This means there are random variations across the environment set $E$, in map layouts and in the existence, location, and names of objects, etc. Consequently, an agent cannot answer questions merely by memorizing games it has seen before.
Because environments are partially observable (i.e., not all necessary information is available at a single turn), an agent must take a sequence of decisions -- analogous to following a search and reasoning procedure -- to gather the required information.
The learning target in \qait is thus not the declarative knowledge $a$ itself, but the \emph{procedure} for arriving at $a$ by collecting evidence.

The main contributions of this work are as follows:
\begin{enumerate}
    \item We introduce a novel MRC dataset, \qait, which focuses on \emph{procedural} knowledge. In it, an agent interacts with an environment to discover the answer to a given question.
    \item We introduce to the MRC domain the practice of generating training data on the fly. We sample training examples from a distribution; hence, an agent is highly unlikely to encounter the same training example more than once. This helps to prevent overfitting and rote memorization.
    \item We evaluate a collection of baseline agents on \qait, including state-of-the-art deep reinforcement learning agents and humans, and discuss limitations of existing approaches.
\end{enumerate}

\section{The \qait Dataset}
\label{section:qait_dataset}

\subsection{Overview}
We make the question answering problem interactive by building text-based games along with relevant question-answer pairs. 
We use TextWorld \citep{cote18textworld} to generate these games.
Each interactive environment is composed of multiple locations with paths connecting them in a randomly drawn graph. 
Several interactable objects are scattered across the locations. 
A player sends text commands to interact with the world, while the game's interpreter only recognizes a small subset of all possible command strings (we call these the valid commands). 
The environment changes state in response to a valid command and returns a string of text feedback describing the change.

The underlying game dynamics arise from a set of objects (e.g., doors) that possess attributes (e.g., doors are openable), and a set of rules (e.g., opening a closed door makes the connected room accessible). The supported attributes are shown in Table~\ref{tab:attributes}, while the rules can be inferred from the list of supported commands (see Appendix~\ref{appd:cmd_list}). Note that player interactions might affect an object's attributes. For instance, cooking a piece of \emph{raw chicken} on the stove with a frying pan makes it edible, transforming it into \emph{fried chicken}.

In each game, the existence of objects, the location of objects, and their names are randomly sampled. Depending on the task, a name can be a made-up word. However, game dynamics are constant across all games -- e.g., there will never be a drinkable heat source.

Text in \qait is generated by the TextWorld engine according to English templates, so it does not express the full variation of natural language. However, taking inspiration from the bAbI tasks \citep{weston2015babi}, we posit that controlled simplifications of natural language are useful for isolating more complex reasoning behaviors.


\begin{table}[t]
    \scriptsize
    \centering
    \begin{tabular}{c|c|c|c|c|c|c|c|c|c}
        \toprule
        & \rot{\textbf{edible}} & \rot{\textbf{drinkable}} & \rot{\textbf{portable}} & \rot{\textbf{openable}} & \rot{\textbf{cuttable}} & \rot{\textbf{sharp}} & \rot{\textbf{heat\_source}} & \rot{\textbf{cookable}} & \rot{\textbf{holder}} \\
        \midrule
        Butter knife &  &  & \checkmark &  &  & \checkmark &  &  &  \\
        Oven &  &  &  & \checkmark &  &  & \checkmark &  & \checkmark \\
        Raw chicken &  &  & \checkmark &  & \checkmark &  &  & \checkmark &  \\
        Fried chicken & \checkmark &  & \checkmark &  & \checkmark &  &  & \checkmark &  \\
        \bottomrule
    \end{tabular}
    \caption{Supported attributes along with examples.}
    \label{tab:attributes}
\end{table}

\subsection{Available Information}
\label{section:available_info}

At every game step, the environment returns an \emph{observation string} describing the information visible to the agent, as well as the \emph{command feedback}, which is text describing the response to the previously issued command.

\textbf{Optional Information:}
Since we have access to the underlying state representation of a generated game, various optional information can be made available.
For instance, it is possible to access the subset of commands that are valid at the current game step.
Other available meta-information includes all objects that exist in the game, plus their locations, attributes, and states. 

During training, one is free to use any optional information to guide the agent's learning, e.g., to shape the rewards. However, at test time, only the \emph{observation string} and the \emph{command feedback} are available.


\subsection{Question Types and Difficulty Levels}
\label{section:questions}
Using the game information described above, we can generate questions with known ground truth answers for any given game.

\subsubsection{Question Types}
\label{section:question_types}
For this initial version of \qait we consider three straightforward question types.

\paragraph{Location:} (``Where is the can of soda?'') Given an object name, the agent must answer with the name of the container that most directly holds the object. This can be either a location, a holder within a location, or the player's inventory. For example, if the can of soda is in a fridge which is in the kitchen, the answer would be ``fridge''.
\paragraph{Existence:} (``Is there a raw egg in the world?'') Given the name of an object, the agent must learn to answer whether the object exists in the game environment $e$.
\paragraph{Attribute:} (``Is \textit{ghargh} edible?'') Given an object name and an attribute, the agent must answer with the value of the given attribute for the given object. Note that all attributes in our dataset are binary-valued. To discourage an agent from simply memorizing attribute values given an object name \citep{anand18blindfold} (e.g., apples are always edible so agents can answer without interaction), we replace object names with unique, randomly drawn made-up words for this question type.

\subsubsection{Difficulty Levels}
\label{section:difficulty_levels}
To better analyze the limitations of learning algorithms and to facilitate curriculum learning approaches, we define two difficulty levels based on the environment layout.

\paragraph{Fixed Map:} The map (location names and layout) is fixed across games. Random objects are distributed across the map in each game. Statistics for this game configuration are shown in Table~\ref{tab:data_stats}.

\paragraph{Random Map:} Both map layouts and objects are randomly sampled in each game.

\begin{table}[h!]
    \footnotesize
    \centering
    \begin{tabular}{r|c|c}
        \toprule
        & Fixed Map & Random Map \\
        \midrule
        \# Locations, $N_r$ & 6 & $N_r \sim \text{Uniform}[2, 12]$ \\
        \midrule
        \# Entities, $N_e$ & \multicolumn{2}{c}{$N_e \sim \text{Uniform}[3 \cdot N_r, 6 \cdot N_r]$}\\
        \midrule
        Actions / Game & 17 & 17 \\
        Modifiers / Game & 18.5 & 17.7 \\
        Objects / Game & 26.7 & 27.5 \\
        \midrule
        \# Obs. Tokens & 93.1 & 89.7 \\
        \bottomrule
    \end{tabular}
    \caption{Statistics of the \qait dataset. Numbers are averaged over 10,000 randomly sampled games.}
    \label{tab:data_stats}
\end{table}

\subsection{Action Space}
\label{section:action_space}

We describe the action space of \qait by splitting it into two subsets: information-gathering actions and question-answering actions.

\paragraph{Information Gathering}
The player generates text commands word by word to navigate through and interact with the environment.
On encountering an object, the player must interact with it to discover its attributes. To succeed, an agent must map the feedback received from the environment, in text, to a useful state representation. This is a form of reading comprehension.

To make the \qait task more tractable, all text commands are triplets of the form \{action, modifier, object\} (e.g., \cmd{open wooden door}). When there is no ambiguity, the environment understands commands without modifiers (e.g., \cmd{eat apple} will result in eating the ``red apple'' provided it is the only apple in the player's inventory).
We list all supported commands in Appendix~\ref{appd:cmd_list}.

Each game provides a set of three lexicons that divide the full vocabulary into actions, modifiers, and objects.
Statistics are shown in Table~\ref{tab:data_stats}.
A model can generate a command at each game step by, e.g., sampling from a probability distribution induced over each lexicon.
This reduces the size of the action space compared to a sequential, free-form setting where a model can pick any vocabulary word at any generation step.

An agent decides when to stop interacting with the environment to answer the question by generating a special \cmd{wait} command \footnote{We call it ``wait'' because when playing multiple games in a batch, batched environment will terminate only when all agents have issued the terminating command. Before that, some agent will wait. This is analogous to paddings in natural language processing tasks.}. 
However, the number of interaction steps is limited: we use 80 steps in all experiments. 
When an agent has exhausted its available steps, the game terminates and the agent is forced to answer the question.

\paragraph{Question Answering}
Currently, all \qait answers are one word. For existence and attribute questions, the answer is either \cmd{yes} or \cmd{no}; for location questions, the answer can be any word in an observation string. 

\subsection{Evaluation Settings and Metrics}
\label{section:eval_metrics}
We evaluate an agent's performance on \qait by its accuracy in answering questions. We propose three distinct settings for the evaluation. 

\paragraph{Solving Training Games:} We use QA accuracy during training, averaged over a window of training time, to evaluate an agent's training performance. We provide 5 training sets for this purpose with [1, 2, 10, 100, 500] games, respectively. Each game in these sets is associated with multiple questions.

\paragraph{Unlimited Games:} We implement a setup where games are randomly generated on the fly during training, rather than selected from a finite set as above. The distribution we draw from is controlled by a few parameters: number of locations, number of objects, type of map, and a random seed. From the \textbf{fixed map} game distribution described in Table~\ref{tab:data_stats}, more than $\num{e40}$ different games can be drawn. This means that a game is unlikely to be seen more than once during training. We expect that only a model with strong generalization capabilities will perform well in this setting. 

\paragraph{Zero-shot Evaluation:} For each game setting and question type, we provide 500 held out games that are never seen during training, each with one question. These are used to benchmark generalization in models in a reproducible manner, no matter the training setting. This set is analogous to the test set used in traditional supervised learning tasks, and can be used in conjunction with any training setting.

\section{Baseline Models}
\label{section:baseline}

\subsection{Random Baseline}
Our simplest baseline does not interact with the environment to answer questions; it samples an answer word uniformly from the QA action space (\cmd{yes} and \cmd{no} for attribute and existence questions; all possible object names in the game for location questions).



\subsection{Human Baseline}
\label{section:human}
We conducted a study with 21 participants to explore how humans perform on \qait in terms of QA accuracy. 
Participants played games they had not seen previously from a set generated by sampling 4 game-question pairs for each question type and difficulty level. The human results presented below always represent an average over 3 participants.

\subsection{\mdqn}

\begin{figure}[t]
    \centering
    \includegraphics[width=0.5\textwidth]{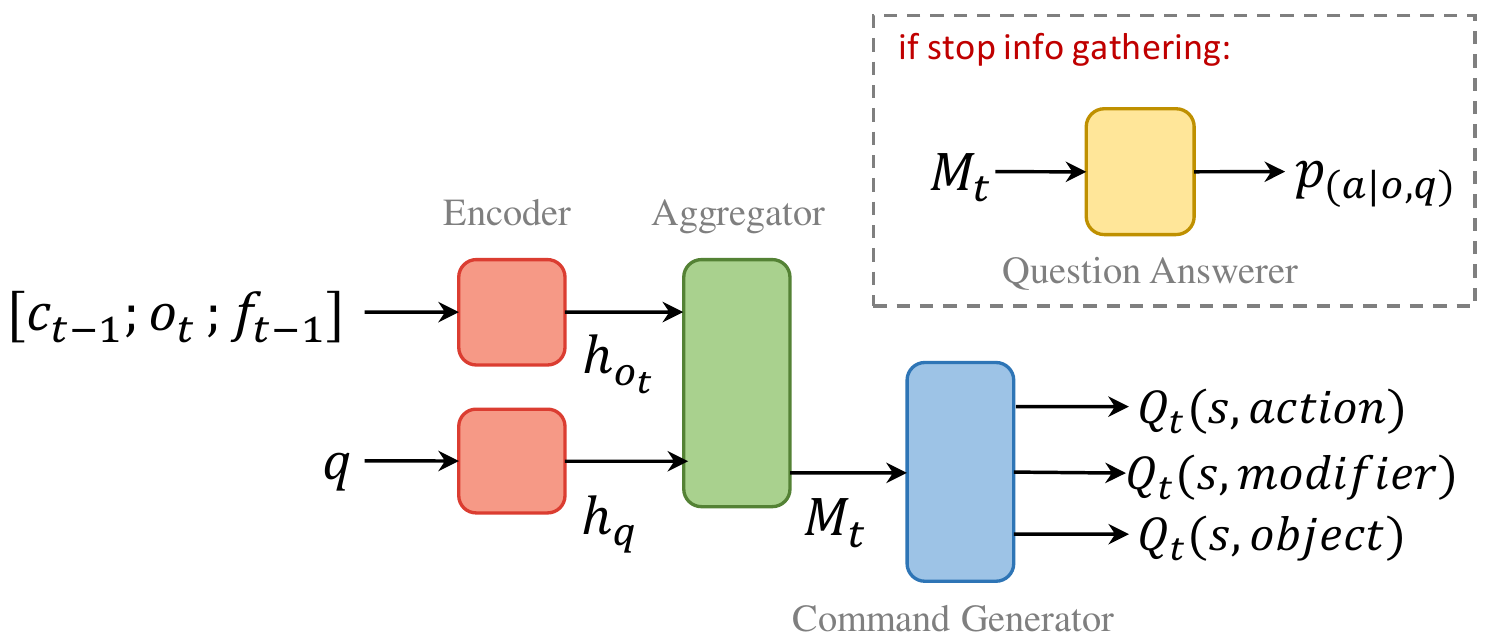}
    \caption{Overall architecture of our baseline agent.}
    \label{fig:model_structure}
\end{figure}

We propose a neural baseline agent, \mdqn, which takes inspiration from the work of \citet{narasimhan15lstmdqn} and \citet{yu18qanet}. 
The agent consists of three main components: an encoder, a command generator, and a question answerer.
More precisely, at game step $t$, the encoder takes observation $o_t$ and question $q$ as input to generate hidden representations.\footnote{We concatenate $o_t$ with the command generated at previous game step and the text feedback returned by the game, as described in Section~\ref{section:available_info}.}
In the information gathering phase, the command generator generates Q-values for all action, modifier, and object words, with rankings of these Q-values used to generate text commands $c_t$.
At any game step, the agent may decide to terminate information gathering and answer the question (or it is forced to do so if it has used up all of its moves).
The question answerer uses the hidden representations at the final information-gathering step to generate a probability distribution over possible answers.

An overview of this architecture is shown in Figure~\ref{fig:model_structure} and full details are given in Appendix~\ref{appd:mdqn}.

\subsubsection{Reward Shaping}
\label{section:reward_shaping}

We design the following two rewards to help \mdqn learn more efficiently; both used for training the command generator.
Note that these rewards are part of the design of \mdqn, but are not used to evaluate its performance. Question answering accuracy is the only evaluation metric for \qait tasks.



\paragraph{Sufficient Information Bonus:} To tackle \qait tasks, an intelligent agent should know when to stop interacting -- it should stop as soon as it has gathered enough information to answer the question correctly. For guiding the agent to learn this behavior, we give an additional reward when the agent stops with sufficient information. Specifically, assuming the agent decides to stop at game step $k$:
\begin{itemize}
    \item Location: reward is 1 if the entity mentioned in the question is a sub-string of $o_k$, otherwise it is 0. This means whenever an agent observes the entity, it has sufficient information to infer the entity's location.

    \item Existence: when the correct answer is \cmd{yes}, a reward of 1 is assigned only if the entity is a sub-string of $o_k$. When the correct answer is \cmd{no}, a reward between 0 and 1 is given. The reward value corresponds to the exploration coverage of the environment, i.e., how many locations the agent has visited, and how many containers have been opened. 

    \item Attribute: we heuristically define a set of conditions to verify each attribute, and reward the agent based on its fulfilment of these conditions. For instance, determining if an object \cmd{X} is \textbf{sharp} corresponds to checking the outcome of a cut command (\cmd{slice}, \cmd{chop}, or \cmd{dice}) while holding the object \cmd{X} and a cuttable food item. If the outcome is successful then the object \cmd{X} is sharp otherwise it is not. Alternatively, if trying to take the object \cmd{X} results in a failure, then we can deduces it is not sharp as all sharp objects are portable. The list of conditions for each attribute used in our experiments is shown in Appendix~\ref{appd:reasoning}.

\end{itemize}

\paragraph{Episodic Discovery Bonus:} Following \citet{yuan18coin}, we use an episodic counting reward to encourage the agent to discover unseen game states. The agent is assigned a positive reward whenever it encounters a new state (in text-based games, states are simply represented as strings):
\begin{center}
$
  r(o_t) = 
      \begin{cases}
          1.0 & \text{if $n(o_t) = 1$,}\\
          0.0 & \text{otherwise},
      \end{cases}
$\\
\end{center}
where $n(\cdot)$ is reset to zero after each episode.


\subsubsection{Training Strategy}
We apply different training strategies for the command generator and the question answerer.

\textbf{Command Generation:} 
Text-based games are sequential decision-making problems that can be described naturally by partially observable Markov decision processes (POMDPs)~\citep{kaelbling98planning}.
We use the Q-Learning \citep{watkins1992qlearning} paradigm to train our agent. 
Specifically, following \citet{mnih2015dqn}, our Q-value function is approximated with a deep neural network.
Beyond vanilla DQN, we also apply several extensions, such as Rainbow \citep{hessel17rainbow}, to our training process.
Details are provided in Section~\ref{section:exp}.

\textbf{Question Answering:} 
During training, we push all question answering transitions (observation strings when interaction stops, question strings, ground-truth answers) into a replay buffer. 
After every 20 game steps, we randomly sample a mini-batch of such transitions from the replay buffer and train the question answerer with supervised learning (e.g., using negative log-likelihood (NLL) loss).

\section{Experimental Results}
\label{section:exp}

In this section, we report experimental results by difficulty levels. 
All random baseline performance values are averaged over 100 different runs.
In the following subsections, we use ``DQN'', ``DDQN'' and ``Rainbow'' to indicate \mdqn trained with vanilla DQN, Double DQN with prioritized experience replay, and Rainbow, respectively. 
Training curves shown in the following figures represent a sliding-window average with a window size of 500. Moreover, each curve is the average of 3 random seeds. For evaluation, we selected the model with the random seed yielding the highest training accuracy to compute its accuracy on the test games. Due to space limitations, we only report some key results here. See Appendix~\ref{appd:result_full} for the full experimental results.

\begin{figure}[h!]
    \centering
    \includegraphics[width=0.5\textwidth]{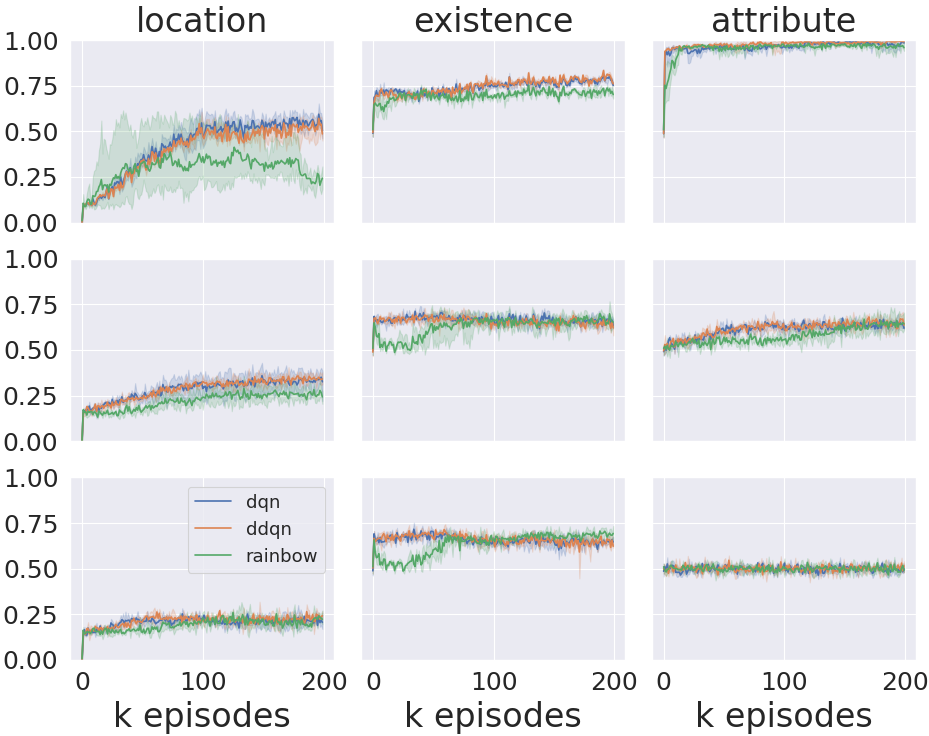}
    \caption{Training accuracy over episodes on \textbf{fixed map} setup. Upper row: 10 games; middle row: 500 games; lower row: unlimited games.}
    \label{fig:fixed_10_500_unlimited}
\end{figure}

\begin{table}[h!]
    \centering
    \scriptsize
    \begin{tabular}{r|c|c|c|c|c|c}
        \toprule
        &  \multicolumn{3}{c|}{Fixed Map} & \multicolumn{3}{c}{Random Map} \\
        \midrule 
        Model & Loc. & Exi. & Att. & Loc. & Exi. & Att.\\
        \midrule 
        \text{Human} & 1.000 & 1.000 & 1.000 & 1.000 & 1.000 & 0.750 \\
        \text{Random} & 0.027 & 0.497 & 0.496  & 0.034 & 0.500 & 0.499\\
        \midrule 
        \multicolumn{7}{c}{10 games}\\
        \midrule
        \text{DQN}      & 0.180 & 0.568 & 0.518 & 0.156 & 0.566 & 0.518 \\
        \text{DDQN}     & 0.188 & 0.566 & 0.516 & 0.142 & 0.606 & 0.500 \\
        \text{Rainbow}  & 0.156 & 0.590 & 0.520 & 0.144 & 0.586 & \textbf{0.530} \\
        \midrule
        \multicolumn{7}{c}{500 games}\\
        \midrule
        \text{DQN}      & 0.224 & 0.674 & \textbf{0.534} & 0.204 & 0.678 & \textbf{0.530} \\
        \text{DDQN}     & 0.218 & 0.626 & 0.508 & 0.222 & 0.656 & 0.486 \\
        \text{Rainbow}  & 0.190 & 0.656 & 0.496 & 0.172 & 0.678 & 0.494 \\
        \midrule
        \multicolumn{7}{c}{unlimited games}\\
        \midrule
        \text{DQN}      & 0.216 & 0.662 & 0.514 & 0.188 & 0.668 & 0.506 \\
        \text{DDQN}     & 0.258 & 0.628 & 0.480 & 0.206 & \textbf{0.694} & 0.482 \\
        \text{Rainbow}  & \textbf{0.280} & \textbf{0.692} & 0.514 & \textbf{0.258} & 0.686 & 0.470 \\
        \bottomrule
    \end{tabular}
    \caption{Agent performance on zero-shot test games when trained on 10 games, 500 games and ``unlimited'' games settings. Note Att. and Exi. are binary questions with expected accuracy of 0.5. }
    \label{tab:zero_shot_test}
\end{table}

\subsection{Fixed Map}
Figure~\ref{fig:fixed_10_500_unlimited} shows the training curves for the neural baseline agents when trained using 10 games, 500 games and the ``unlimited'' games settings.
Table~\ref{tab:zero_shot_test} reports their zero-shot test performance.

From Figure~\ref{fig:fixed_10_500_unlimited}, we observe that when training data size is small (e.g., 10 games), our baseline agent trained with all the three RL methods successfully master the training games.
Vanilla DQN and DDQN are particularly strong at memorizing the training games.
When training on more games (e.g., 500 games and unlimited games), in which case memorization is more difficult, Rainbow agents start to show its superiority --- it has similar accuracy as the other two methods, and even outperforms them in existence question type.

From Table~\ref{tab:zero_shot_test} we see similar observation, when trained on 10 games and 500 games, DQN and DDQN performs better on test games but on the unlimited games setting, rainbow agent performs as good as them, and sometimes even better. 
We can also observe that our agents fail to generalize on attribute questions. 
In unlimited games setting as shown in Figure~\ref{fig:fixed_10_500_unlimited}, all three agents produce an accuracy of 0.5; in zero-shot test as shown in Table~\ref{tab:zero_shot_test}, no agent performs significantly better than random.
This suggests the agents memorize game-question-answer triples when data size is small, and fail to do so in unlimited games setting.
This can also be observed in Appendix~\ref{appd:result_full}, where in attribute question experiments, the training accuracy is high, and sufficient information bonus is low (even close to 0).

\begin{figure}[t]
    \centering
    \includegraphics[width=0.5\textwidth]{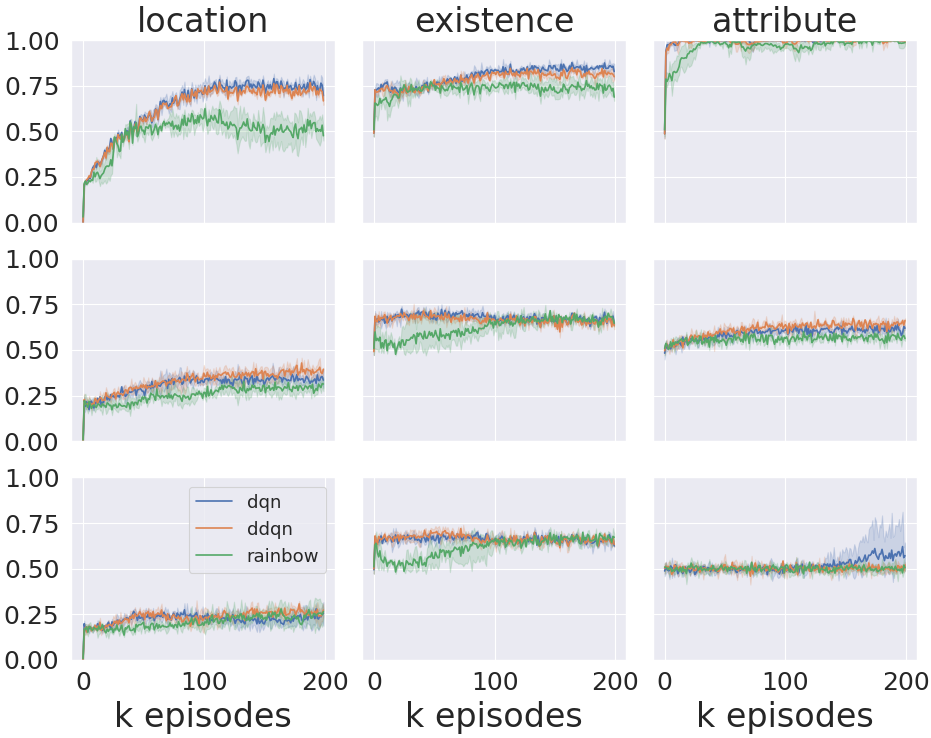}
    \caption{Training accuracy on the \textbf{random map} setup. Upper row: 10 games; middle row: 500 games; lower row: unlimited games.}
    \label{fig:rand_10_500_unlimited}
\end{figure}

\subsection{Random Map}

Figure~\ref{fig:rand_10_500_unlimited} shows the training curves for the neural baseline agents when trained using 10 games, 500 games and ``unlimited'' games settings.
The trends of our agents' performance on random map games are consistent with on fixed map games.
However, because there exist easier games (as listed in Table~\ref{tab:data_stats}, number of rooms is sampled between 2 and 12), agents show better training performance in such setting than \textbf{fixed map} setting in general.

Interestingly, we observe one of the DQN agent starts to learn in the unlimited games, attribute question setting. 
This may be because in games with smaller map size and less objects, there is a higher chance to accomplish some sub-tasks (e.g., it is easier to find an object when there are less rooms), and the agent learn such skills and apply them to similar tasks.
Unfortunately, as shown in Table~\ref{tab:zero_shot_test} that agent does not perform significantly better than random on test set. 
We expect with more training episodes, the agent can have a better generalization performance.

\subsection{Question Answering Given Sufficient Information}

\begin{table}
\centering
\scriptsize
\begin{tabular}{r|c|c}
    \toprule
    Model &  Fixed Map & Random Map \\
    \midrule
    \text{Random} & 14.7  & 16.5\\
    \midrule
    \multicolumn{3}{c}{10 games}\\
    \midrule
    \text{DQN}      & 95.7  & 97.5\\
    \text{DDQN}     & 90.4  & 92.2\\
    \text{Rainbow}  & 91.8  & 84.7\\
    \midrule
    \multicolumn{3}{c}{500 games}\\
    \midrule
    \text{DQN}      & 91.8  & 94.4\\
    \text{DDQN}     & 95.6  & 90.2\\
    \text{Rainbow}  & 96.9  & 96.6\\
    \midrule
    \multicolumn{3}{c}{unlimited games}\\
    \midrule
    \text{DQN}      & 100.0   & 100.0\\
    \text{DDQN}     & 100.0  & 100.0\\
    \text{Rainbow}  & 100.0   & 100.0\\
    \bottomrule
\end{tabular}
\caption{Test performance given sufficient information.}
\label{tab:suff_info}
\end{table}


The challenge in \qait is learning the interactive procedure for arriving at a state with the information needed to answer the question. 
We conduct the following experiments on \textbf{location} questions to investigate this challenge.

Based on the results in Table~\ref{tab:zero_shot_test}, we compute an agent's test accuracy \emph{only if} it has obtained sufficient information -- i.e., when the sufficient information bonus is 1. Results shown in Table~\ref{tab:suff_info} support our assumption that the QA module can learn (and generalize) effectively to answer given sufficient information.
Similarly, experiments show that when objects being asked about are in the current observation, the random baseline's performance goes up significantly as well.
We report our baseline agents' question answering accuracy and sufficient information bonuses on all experiment settings in Appendix~\ref{appd:result_full}.

\subsection{Full Information Setup}

To reframe the \qait games as a standard MRC task, we also designed an experimental setting that eliminates the need to gather information interactively.
From a heuristic trajectory through the game environment that is guaranteed to observe sufficient information for $q$, we concatenate all observations into a static ``document'' $d$ to build a $\{d, q, a\}$ triplet.
A model then uses this fully observed document as input to answer the question.
We split this data into training, validation, and test sets and follow the evaluation protocol for standard supervised MRC tasks.
We take an off-the-shelf MRC model, Match-LSTM \cite{wang16matchlstm}, trained with negative log-likelihood loss as a baseline.

Unsurprisingly, Match-LSTM does fairly well on all 3 question types (86.4, 89.9 and 93.2 test accuracy on location, existence, and attribute questions, respectively). This implies that without the need to interact with the environment for information gathering, the task is simple enough that a word-matching model can answer questions with high accuracy.

\section{Related Work}
\label{sec:related_work}

\subsection{MRC Datasets}
Many large-scale machine reading comprehension and question answering datasets have been proposed recently.
The datasets of \citet{rajpurkar16squad, trischler16newsqa} contain crowd-sourced questions based on single documents from Wikipedia and CNN news, respectively. 
\citet{nguyen16msmarco, joshi17triviaqa, dunn17searchqa, clark18arc, kwiatkowski19naturalquestions} present question-answering corpora harvested from information retrieval systems, often containing multiple supporting documents for each question. This means a model must sift through a larger quantity of information and possibly reconcile competing viewpoints.
\citet{berant13webquestions, weibl17qangaroo, talmor18complexwebquestion} propose to leverage knowledge bases to generate question-answer pairs. \citet{yang18hotpot} focuses on questions that require multi-hop reasoning to answer, by building questions compositionally. \citet{reddy18coqa, choi18quac} explore conversational question answering, in which a full understanding of the question depends on the conversation's history.

Most of these datasets focus on declarative knowledge and are static, with all information fully observable to a model. We contend that this setup, unlike \qait, encourages word matching. Supporting this contention, several studies highlight empirically that existing MRC tasks require little comprehension or reasoning.
In \citet{Rychalska2018}, it was shown that a question's main verb exerts almost no influence on the answer prediction: in over 90\% of examined cases, swapping verbs for their antonyms does not change a system's decision. \citet{jia2017adversarial} show the accuracy of neural models drops from an average of 75\% $F_1$ score to 36\% $F_1$ when they manually insert adversarial sentences into SQuAD.

\subsection{Interactive Environments}

Several embodied or visual question answering datasets have been presented recently to address some of the problems of interest in our work, such as those of \citet{brodeur17home, das17embodiedqa, gordon17iqa}. In contrast with these, our purely text-based environment circumvents challenges inherent to modelling interactions between separate data modalities.
Furthermore, most visual question answering environments only support navigating and moving the camera as interactions.
In text-based environments, however, it is relatively cheap to build worlds with complex interactions. This is because text enables us to model interactions abstractly without the need for, e.g., a costly physics engine.

Closely related to \qait is BabyAI \citep{boisvert18babyai}. BabyAI is a gridworld environment that also features constrained language for generating simple home-based scenarios (i.e., instructions). However, observations and actions in BabyAI are not text-based.
World of Bits \citep{shi17worldofbits} is a platform for training agents to interact with the internet to accomplish tasks like flight booking. Agents generally do not need to gather information in World of Bits, and the focus is on accomplishing tasks rather than answering questions.

\subsection{Information Seeking}
Information seeking behavior is an important capacity of intelligent systems that has been discussed for many years.
\citet{kuhlthau04seeking} propose a holistic view of information search as a six-stage process.
\citet{schmidhuber10theory} discusses the connection between information seeking and formal notions of fun, creativity, and intrinsic motivation.
\citet{das18dynamickg} propose a model that continuously determines all entities' locations during reading
and dynamically updates the associated representations in a knowledge graph. 
\citet{bachman16infoseeking} propose a collection of tasks and neural methods for learning to gathering information efficiently in an environment.



To our knowledge, we are the first to consider interactive information-seeking tasks for question answering in worlds with complex dynamics.
The \qait task was designed such that simple word matching methods do not apply, while more human-like information seeking models are encouraged.

\section{Discussion and Future Work}
\label{section:discussion}



\paragraph{Monitoring Information Seeking:}
In \qait, the only evaluation metric is question answering accuracy.
However, the sufficient information bonus described in Section~\ref{section:reward_shaping} is helpful for monitoring agents' ability to gather relevant information. 
We report its value for all experiments in Appendix~\ref{appd:result_full}. 
We observe that while the baseline agents can reach a training accuracy of 100\% for answering attribute questions when trained on a few games, the sufficient information bonus is close to 0. 
This is a clear indication that the agent overfits to the question-answer mapping of the games rather than learning how to gather useful information. 
This aligns with our observation that the agent does not perform better than random on the unlimited games setting, because it fails to gather the needed information.

\paragraph{Challenges in \qait:}
\qait focuses on learning procedural knowledge from interactive environments, so it is natural to use deep RL methods to tackle it.
Experiments suggest the dataset presents a major challenge for existing systems, including Rainbow, which set the state of the art on Atari games. 
As a simplified and controllable text-based environment, \qait can drive research in both the RL and language communities, especially where they intersect.
Until recently, the RL community focused mainly on solving single environments (i.e., training and testing on the same game). Now, we see a shift towards solving multiple games and testing for generalization \citep{cobbe18coinrun, justesen18generalization}.
We believe \qait serves this purpose.

\paragraph{Templated Language:}
As \qait is based on TextWorld, it has the obvious limitation of using templated English.
However, TextWorld provides approximately 500 \textit{human-written} templates for describing rooms and objects, so some textual diversity exists, and since game narratives are generated compositionally, this diversity increases along with the complexity of a game.
We believe simplified and controlled text environments offer a bridge to full natural language, on which we can isolate the learning of useful behaviors like information seeking and command generation.
Nevertheless, it would be interesting to further diversify the language in \qait, for instance by having human writers paraphrase questions.

\paragraph{Future Work:}
Based on our present efforts to tackle \qait, we propose the following directions for future work.

A \textbf{structured memory} (e.g., a dynamic knowledge graph as proposed in \citet{das18dynamickg, ammanabrolu2019playing}) could be helpful for explicitly memorizing the places and objects that an agent has observed. This is especially useful when an agent must revisit a location or object 
or should avoid doing so.

Likewise, a variety of \textbf{external knowledge} could be leveraged by agents. For instance, incorporating a pretrained language model could improve command generation by imparting knowledge of word and object affordances. In recent work, \citet{hausknecht19nail} show that pretrained modules together with handcrafted sub-policies help in solving text-based games, while \citet{yin2019cooking} use BERT~\cite{devlin18bert} to inject `weak common sense' into agents for text-based games. \citet{ammanabrolu2019transfer} show that knowledge graphs and their associated neural encodings can be used as a medium for domain transfer across text-based games.



In finite game settings we observed significant overfitting, especially for attribute questions -- as shown in Appendix~\ref{appd:result_full}, our agent achieves high QA accuracy but low sufficient information bonus on the single-game setting. 
Sometimes attributes require long procedures to verify, and thus, we believe that denser rewards would help with this problem. One possible solution is to provide \textbf{intermediate rewards} whenever the agent achieves a sub-task.

\section*{Acknowledgments}
The authors thank Romain Laroche, R\'emi Tachet des Combes, Matthew Hausknecht, Philip Bachman, and Layla El Asri for  insightful ideas and discussions.
We thank Tavian Barnes, Wendy Tay, and Emery Fine for their work on the TextWorld framework.
We also thank the anonymous EMNLP reviewers for their helpful feedback and suggestions.




\bibliography{biblio}{}
\bibliographystyle{apalike}

\clearpage
\appendix

\section{Details of \mdqn}
\label{appd:mdqn}

\subsection*{Notations}

In this section, we use \textit{game step} $t$ to denote one round of interaction between an agent with the \qait environment.
We use $o_t$ to denote text observation at game step $t$, and $q$ to denote question text.
We use $L$ to refer to a linear transformation.
Brackets $[\cdot;\cdot]$ denote vector concatenation.

\subsection{Encoder}

We use a transformer-based text encoder, which consists of an embedding layer, two stacks of transformer blocks (denoted as encoder transformer blocks and aggregation transformer blocks), and an attention layer.

In the embedding layer, we aggregate both word- and character-level information to produce a vector for each token in text.
Specifically, word embeddings are initialized by the 300-dimensional fastText \cite{mikolov18fasttext} word vectors trained on Common Crawl (600B tokens), they are fixed during training.
Character level embedding vectors are initialized with 32-dimensional random vectors. 
A convolutional layer with 64 kernels of size 5 is then used to aggregate the sequence of characters. 
We use a max pooling layer on the character dimension, then a multi-layer perceptron (MLP) of output size 64 is used to aggregate the concatenation of word- and character-level representations.
Highway network \cite{srivastava15highway} is applied on top of this MLP.
The resulting vectors are used as input to the encoding transformer blocks.

Each encoding transformer block consists of a stack of convolutional layers, a self-attention layer, and an MLP. 
In which, each convolutional layer has 64 filters, each kernel's size is 7, there are 2 such convolutional layers that share weights.
In the self-attention layer, we use a block hidden size of 64, as well as a single head attention mechanism.
Layernorm and dropout are applied after each component inside the block. 
We add positional encoding into each block's input. 
We use one layer of such an encoding block. 

At a game step $t$, the encoder processes text observation $o_t$ and question $q$, context aware encoding $h_{o_t} \in \mathbb{R}^{L^{o_t} \times H_1}$ and $h_q \in \mathbb{R}^{L^{q} \times H_1}$ are generated, where $L^{o_t}$ and $L^{q}$ denote number of tokens in $o_t$ and $q$ respectively, $H_1$ is 64.
Following \cite{yu18qanet}, we use an context-query attention layer to aggregate the two representations $h_{o_t}$ and $h_q$.

Specifically, the attention layer first uses two MLPs to convert both $h_{o_t}$ and $h_q$ into the same space, the resulting tensors are denoted as $h_{o_t}' \in \mathbb{R}^{L^{o_t} \times H_2}$ and $h_q' \in \mathbb{R}^{L^{q} \times H_2}$, in which $H_2$ is 64.

Then, a tri-linear similarity function is used to compute the similarities between each pair of $h_{o_t}'$ and $h_q'$ items: 

\begin{equation}
S = W[h_{o_t}'; h_q'; h_{o_t}' \odot h_q'],
\end{equation}
where $\odot$ indicates element-wise multiplication, $W$ is trainable parameters of size 64.

Softmax of the resulting similarity matrix $S$ along both dimensions are computed, this produces $S^A$ and $S^B$. 
Information in the two representations are then aggregated by:

\begin{equation}
\begin{aligned}
h_{oq} &= [h_{o_t}'; P; h_{o_t}'\odot P; h_{o_t}' \odot Q], \\
P &= S_q h_q'^{\top}, \\
Q &= S_q S_{o_t}^{\top} h_{o_t}'^{\top}, \\
\end{aligned}
\end{equation}

where $h_{oq}$ is aggregated observation representation.

On top of the attention layer, a stack of aggregation transformer blocks is used to further map the observation representations to action representations and answer representations. 
The structure of aggregation transformer blocks are the same as the encoder transformer blocks, except the kernel size of convolutional layer is 5, and the number of blocks is 3.

Let $M_t \in \mathbb{R}^{L^{o_t} \times H_3}$ denote the output of the stack of aggregation transformer blocks, where $H_3$ is 64.

\subsection{Command Generator}

The command generator takes the hidden representations $M_t$ as input, it estimates Q-values for all action, modifier, and object words, respectively. 
It consists of a shared Multi-layer Perceptron (MLP) and three MLPs for each of the components:
\begin{equation}
\label{eqn:command_generator}
\begin{aligned}
R_t = \text{ReLU}(&L_{\text{shared}}(\text{mean}(M_t)), \\
Q_{t, action} = &L_{\text{action}}(R_t), \\
Q_{t, modifier} = &L_{\text{modifier}}(R_t), \\
Q_{t, object} = &L_{\text{object}}(R_t). \\
\end{aligned}
\end{equation}

In which, the output size of $L_{\text{shared}}$ is 64; the dimensionalities of the other 3 MLPs are depending on the number of the amount of action, modifier and object words available, respectively.
The overall Q-value is the sum of the three components:
\begin{equation}
Q_{t} = Q_{t, action} + Q_{t, modifier} + Q_{t, object}. \\
\end{equation}

\subsection{Question Answerer}
Similar to \citep{yu18qanet}, we append an extra stacks of aggregation transformer blocks on top of the aggregation transformer blocks to compute answer positions:

\begin{equation}
\begin{aligned}
U &= \text{ReLU}(L_0\lbrack M_t; M_t'\rbrack). \\
\beta &= \smx(L_1(U)). \\
\end{aligned}
\end{equation}

In which $M_t' \in \mathbb{R}^{L^{o_t} \times H_3}$ is output of the extra transformer stack,
$L_0$, $L_1$ are trainable parameters with output size 64 and 1, respectively.

For location questions, the agent outputs $\beta$ as the probability distribution of each word in observation $o_t$ being the answer of the question.

For binary classification questions, we apply an MLP, which takes weighted sum of matching representations as input, to compute a probability distribution $p(y)$ over both possible answers:

\begin{equation}
\begin{aligned}
D &= \sum_i (\beta^i \cdot M_t') ,\\
p(y) &= \smx(L_{4}(\tanh(L_{3}(D))).
\end{aligned}
\end{equation}
Output size of $L_3$ and $L_4$ are 64 and 2, respectively.

\subsection{Deep Q-Learning}
In a text-based game, an agent takes an action $a$\footnote{In our case, $a$ is a triplet contains \{action, modifier, object\} as described in Section~\ref{section:action_space}.} in state $s$ by consulting a state-action value function $Q(s, a)$, this value function is as a measure of the action's expected long-term reward. Q-Learning helps the agent to learn an optimal $Q(s, a)$ value function. The agent starts from a random Q-function, it gradually updates its Q-values by interacting with environment, and obtaining rewards. Following \citet{mnih2015dqn}, the Q-value function is approximated with a deep neural network.

We make use of a replay buffer. During playing the game, we cache all transitions into the replay buffer without updating the parameters. We periodically sample a random batch of transitions from the replay buffer. In each transition, we update the parameters $\theta$ to reduce the discrepancy between the predicted value of current state $Q(s_t, a_t)$ and the expected Q-value given the reward $r_t$ and the value of next state $\max_{a}Q(s_{t+1}, a)$.

We minimize the temporal difference (TD) error, $\delta$:
\begin{equation}
\label{eqn:td_cmd}
\delta = Q(s_t, a_t) - (r_t + \gamma \max_{a}Q(s_{t+1}, a)),
\end{equation}
in which, $\gamma$ indicates the discount factor. 
Following the common practice, we use the Huber loss to minimize the TD error.
For a randomly sampled batch with batch size $B$, we minimize:

\begin{equation}
\begin{aligned}
\mathcal{L} &= \frac{1}{|B|}\sum{\mathcal{L}(\delta)}, \\
\text{where } \mathcal{L}(\delta) &=  \begin{cases}
                                        \frac{1}{2} \delta^2 & \text{for $|\delta| \leq 1$,}\\
                                        |\delta| - \frac{1}{2} & \text{otherwise.}
                                    \end{cases}
\end{aligned}
\end{equation}

As described in Section~\ref{section:reward_shaping}, we design the sufficient information bonus to teach an agent to stop as soon as it has gathered enough information to answer the question.
Therefore we assign this reward at the game step where the agent generates \cmd{wait} command (or it is forced to stop).

It is worth mentioning that for attribute type questions (considerably the most difficult question type in \qait, where the training signal is very sparse), we provide extra rewards to help \mdqn to learn.

Specifically, we take a reward similar to as used in location questions: 1.0 if the agent has observed the object mentioned in the question. 
we also use a reward similar to as used in existence questions: the agent is rewarded by the coverage of its exploration.
The two extra rewards are finally added onto the sufficient information bonus for attribute question, both with coefficient of 0.1.

\section{Implementation Details}
During training with vanilla DQN, we use a replay memory of size 500,000. 
We use $\epsilon$-greedy, where the value of $\epsilon$ anneals from 1.0 to 0.1 within 100,000 episodes.
We start updating parameters after 1,000 episodes of playing.
We update our network after every 20 game steps.
During updating, we use a mini-batch of size 64. 
We use \emph{Adam} \citep{kingma14adam} as the step rule for optimization, The learning rate is set to 0.00025. 

When our agent is trained with Rainbow algorithm, we follow \citet{hessel17rainbow} on most of the hyper-parameter settings.
The four MLPs $L_{\text{shared}}$, $L_{\text{action}}$, $L_{\text{modifier}}$ and $L_{\text{object}}$ as described in Eqn.~\ref{eqn:command_generator} are Noisy Nets layers \citep{fortunato17noisynets} when the agent is trained in Rainbow setting. 
Detailed hyper-parameter setting of our Rainbow agent are shown in Table~\ref{tab:rainbow_hyper_parameter}.

\begin{table}[h!]
    \centering
    \small
    \begin{tabular}{r|c}
        \toprule
        Parameter & Value \\
        \midrule 
        \text{Exploration} $\epsilon$ & 0 \\
        \text{Noisy Nets} $\sigma_0$ & 0.5 \\
        \text{Target Network Period} & 1000 episodes \\
        \text{Multi-step returns} $n$ & $n \sim \text{Uniform}[1, 3]$ \\
        \text{Distributional atoms} & 51 \\
        \text{Distributional min/max values} & [-10, 10] \\
        \bottomrule
    \end{tabular}
    \caption{Hyper-parameter setup for rainbow agent.}
    \label{tab:rainbow_hyper_parameter}
\end{table}
The model is implemented using PyTorch \citep{paszke17automatic}.

\section{Supported Text Commands}
\label{appd:cmd_list}
All supported text commands are listed in Table~\ref{tab:cmd_list}.

\begin{table*}[h!]
    \centering
    \small
    \begin{tabular}{l|l}
        \toprule
        Command & Description \\
        \midrule 
        \cmd{look}	                    & describe the current location \\
        \cmd{inventory}	                & display the player's inventory \\
        \cmd{go $\langle$dir$\rangle$}	& move the player to north, east, south, or west \\
        \cmd{examine ...}	            & examine something more closely \\
        \cmd{open ...}	                & open a door or a container \\
        \cmd{close ...}	                & close a door or a container \\
        \cmd{eat ...}	                & eat edible object \\
        \cmd{drink ...}	                & drink drinkable object \\
        \cmd{drop ...}	                & drop an object on the floor \\
        \cmd{take ...}	                & take an object from the floor, a container, or a supporter \\
        \cmd{put ...}	                & put an object onto a supporter (supporter must be present at the location) \\
        \cmd{insert ...}	            & insert an object into a container (container must be present at the location) \\
        \cmd{cook ...}	                & cook an object (heat source must be present at the location) \\
        \cmd{slice ...}	                & slice cuttable object (a sharp object must be in the player's inventory) \\
        \cmd{chop ...}	                & chop cuttable object (a sharp object must be in the player's inventory \\
        \cmd{dice ...}	                & dice cuttable object (a sharp object must be in the player's inventory) \\
        \cmd{wait}	                    & stop interaction \\
        \bottomrule
    \end{tabular}
    \caption{Supported command list.}
    \label{tab:cmd_list}
\end{table*}

\section{Heuristic Conditions for Attribute Questions}
\label{appd:reasoning}
Here, we derived some heuristic conditions to determine when an agent has gathered enough information to answer a given attribute question. Those conditions are used as part of the reward shaping for our proposed agent (Section~\ref{section:reward_shaping}). In Table~\ref{tab:reasoning}, for each attribute we list all the commands for which their outcome (pass or fail) gives enough information to answer the question correctly. Also, in order for a command's outcome to be informative, each command needs to be executed while some state conditions hold. For example, to determine if an object is indeed a \textbf{heat\_source}, the agent needs to try to cook something that is cookable and uncooked while standing next to the given object.

\begin{table*}[h!]
  \centering
  \small
  \begin{tabular}{l|l|l|c|c|l}
    \toprule
    Attribute & Command & State & Pass & Fail & Explanation \\
    \midrule 
    \multirow{4}{*}{\textbf{sharp}} & \multirow{3}{*}{\cmd{cut} \textit{cuttable}} & holding (\textit{cuttable}) & \multirow{3}{*}{1} & \multirow{3}{*}{1} &
    Trying to cut something cuttable \\
          &                             & \& uncut (\textit{cuttable}) & & & that hasn't been cut yet \\
          &                             & \& holding (object) & & & while holding the object. \\
    \cmidrule(){2-6}
     & \cmd{take} object & reachable(object) & 0 & 1 & 
    Sharp objects should be portable. \\
    \midrule
    \multirow{3}{*}{\textbf{cuttable}} & \multirow{2}{*}{\cmd{cut} object} & holding (object) & \multirow{2}{*}{1} & \multirow{2}{*}{1} & Trying to cut the object while holding \\
    & & \& holding (\textit{sharp}) & & & something sharp. \\
    \cmidrule(){2-6}
    & \cmd{take} object & reachable (object) & 0 & 1 & Cuttable object should be portable. \\
    \midrule
    \multirow{2}{*}{\textbf{edible}} & \cmd{eat} object & holding (object) & 1 & 1 & 
    Trying to eat the object. \\
    \cmidrule(){2-6}
    & \cmd{take} object & reachable (object) & 0 & 1 & 
    Edible objects should be portable. \\
    \midrule
    \multirow{2}{*}{\textbf{drinkable}} & \cmd{drink} object & holding (object) & 1 & 1 & 
    Trying to drink the object. \\
    \cmidrule(){2-6}
     & \cmd{take} object & reachable (object) & 0 & 1 & 
    Drinkable objects should be portable. \\
    \midrule
    \multirow{3}{*}{\textbf{holder}} & \multicolumn{1}{c|}{\multirow{2}{*}{\textbf{--}}} & on (\textit{portable}, object) & 1 & 0 & Observing object(s) on a supporter. \\
    \cmidrule(){3-6}
    & & in (\textit{portable}, object) & 1 & 0 & Observing object(s) inside a container. \\
    \cmidrule(){2-6}
     & \cmd{take} object & reachable (object) & 1 & 0 & Holder objects should not be portable. \\
    \midrule
    \multirow{2}{*}{\textbf{portable}} & \multicolumn{1}{c|}{\textbf{--}} & holding (object) & 1 & 0 & 
    Holding the object means it is portable. \\
    \cmidrule(){2-6}
     & \cmd{take} object & reachable (object) & 1 & 1 & 
    Portable objects can be taken. \\
    \midrule
    \multirow{4}{*}{\textbf{heat\_source}} & \multirow{3}{*}{\cmd{cook} \textit{cookable}} & holding (\textit{cookable}) & \multirow{3}{*}{1} & \multirow{3}{*}{1} & 
    Trying to cook something cookable \\
              &                   & \& uncooked (\textit{cookable}) & & & that hasn't been cooked yet\\
              &                   & \& reachable (object) & & & while being next to the object.\\
    \cmidrule(l){2-6}
     & \cmd{take} object & reachable (object) & 1 & 0 & 
    Heat source objects should not be portable. \\
    \midrule 
    \multirow{3}{*}{\textbf{cookable}} & \multirow{2}{*}{\cmd{cook} object} & holding (object) & \multirow{2}{*}{1} & \multirow{2}{*}{1} &
    Trying to cook the object \\
          &                             & \& reachable (\textit{heat\_source}) & & & while being next to a heat source. \\
    \cmidrule(){2-6}
     & \cmd{take} object & reachable(object) & 0 & 1 & 
    Cookable objects should be portable. \\
    \midrule
    \multirow{4}{*}{\textbf{openable}} & \multirow{2}{*}{\cmd{open} object} & reachable (object) & \multirow{2}{*}{1} & \multirow{2}{*}{1} & \multirow{2}{*}{Trying to open the closed object.} \\
    & & \& closed (object) & & & \\
    \cmidrule(){2-6}
     & \multirow{2}{*}{\cmd{close} object} & reachable (object) & \multirow{2}{*}{1} & \multirow{2}{*}{1} & \multirow{2}{*}{Trying to close the open object.} \\
    & & \& open (object) & & & \\
    \bottomrule
  \end{tabular}
  \caption{
    Heuristic conditions for determining whether the agent has enough information to answer a given attribute question. We use ``object'' to refer to the object mentioned in the question. Words in italics represents placeholder that can be replaced by any object from the environment that has the appropriate attribute (e.g. carrot could be used as a \textit{cuttable}). The columns Pass and Fail represent how much reward the agent will receive given the corresponding command's outcome (resp. success or failure). NB: \cmd{cut} can mean any of the following commands: \cmd{slice}, \cmd{dice}, or \cmd{chop} 
  }
  \label{tab:reasoning}
\end{table*}

\section{Full results}
\label{appd:result_full}
We provide full results of our agents on \textbf{fixed map} games in Table~\ref{tab:fixed_map_full}, and provide full results of our agents on \textbf{random map} games in Table~\ref{tab:rand_map_full}. To help investigating the generalizability of the sufficient information bonus we used in our proposed agent, we also report the rewards during both training and test phases. Note during test phase, we do not update parameters with the rewards.

\clearpage
\begin{table*}[h!]
    \centering
    \small
    \begin{tabular}{r|c|c|c|c|c|c}
        \toprule
        & \multicolumn{2}{c|}{Location} & \multicolumn{2}{c|}{Existence} & \multicolumn{2}{c}{Attribute}\\
        Model & Train & Test & Train & Test & Train & Test \\
        \midrule
        \text{Human} & -- & 1.000 & -- & 1.000 & -- & 1.000 \\
        \text{Random} & -- & 0.027 & -- & 0.497 & -- & 0.496 \\
        \midrule 
        \multicolumn{7}{c}{1 game}\\
        \midrule
        \text{DQN} & 0.972\color{blue}(0.972) & 0.122\color{blue}(0.160) & 1.000\color{blue}(0.881) & 0.628\color{blue}(0.124) & 1.000\color{blue}(0.049) & 0.500\color{blue}(0.035) \\
        \text{DDQN} & 0.960\color{blue}(0.960) & 0.156\color{blue}(0.178) & 1.000\color{blue}(0.647) & 0.624\color{blue}(0.148) & 1.000\color{blue}(0.023) & 0.498\color{blue}(0.033) \\
        \text{Rainbow} & 0.562\color{blue}(0.562) & 0.164\color{blue}(0.178) & 1.000\color{blue}(0.187) & 0.616\color{blue}(0.083) & 1.000\color{blue}(0.049) & 0.516\color{blue}(0.039) \\
        \midrule
        \multicolumn{7}{c}{2 games}\\
        \midrule
        \text{DQN} & 0.698\color{blue}(0.698) & 0.168\color{blue}(0.182) & 0.948\color{blue}(0.700) & 0.574\color{blue}(0.136) & 1.000\color{blue}(0.011) & 0.510\color{blue}(0.028) \\
        \text{DDQN} & 0.702\color{blue}(0.702) & 0.172\color{blue}(0.178) & 0.882\color{blue}(0.571) & 0.550\color{blue}(0.109) & 1.000\color{blue}(0.098) & 0.508\color{blue}(0.036) \\
        \text{Rainbow} & 0.734\color{blue}(0.734) & 0.160\color{blue}(0.168) & 0.878\color{blue}(0.287) & 0.616\color{blue}(0.085) & 1.000\color{blue}(0.030) & 0.524\color{blue}(0.022) \\
        \midrule
        \multicolumn{7}{c}{10 games}\\
        \midrule
        \text{DQN} & 0.654\color{blue}(0.654) & 0.180\color{blue}(0.188) & 0.822\color{blue}(0.390) & 0.568\color{blue}(0.156) & 1.000\color{blue}(0.055) & 0.518\color{blue}(0.030) \\
        \text{DDQN} & 0.608\color{blue}(0.608) & 0.188\color{blue}(0.208) & 0.842\color{blue}(0.479) & 0.566\color{blue}(0.128) & 1.000\color{blue}(0.064) & 0.516\color{blue}(0.036) \\
        \text{Rainbow} & 0.616\color{blue}(0.616) & 0.156\color{blue}(0.170) & 0.768\color{blue}(0.266) & 0.590\color{blue}(0.131) & 0.998\color{blue}(0.059) & 0.520\color{blue}(0.023) \\
        \midrule
        \multicolumn{7}{c}{100 games}\\
        \midrule
        \text{DQN} & 0.498\color{blue}(0.498) & 0.194\color{blue}(0.206) & 0.756\color{blue}(0.139) & 0.614\color{blue}(0.160) & 0.838\color{blue}(0.019) & 0.498\color{blue}(0.014) \\
        \text{DDQN} & 0.456\color{blue}(0.458) & 0.168\color{blue}(0.196) & 0.768\color{blue}(0.134) & 0.650\color{blue}(0.216) & 0.878\color{blue}(0.020) & 0.528\color{blue}(0.017) \\
        \text{Rainbow} & 0.340\color{blue}(0.340) & 0.156\color{blue}(0.160) & 0.762\color{blue}(0.129) & 0.602\color{blue}(0.207) & 0.924\color{blue}(0.044) & 0.524\color{blue}(0.022) \\
        \midrule
        \multicolumn{7}{c}{500 games}\\
        \midrule
        \text{DQN} & 0.430\color{blue}(0.430) & 0.224\color{blue}(0.244) & 0.742\color{blue}(0.136) & 0.674\color{blue}(0.279) & 0.700\color{blue}(0.015) & \textbf{0.534}\color{blue}(0.014) \\
        \text{DDQN} & 0.406\color{blue}(0.406) & 0.218\color{blue}(0.228) & 0.734\color{blue}(0.173) & 0.626\color{blue}(0.213) & 0.714\color{blue}(0.021) & 0.508\color{blue}(0.026) \\
        \text{Rainbow} & 0.358\color{blue}(0.358) & 0.190\color{blue}(0.196) & 0.768\color{blue}(0.187) & 0.656\color{blue}(0.207) & 0.736\color{blue}(0.032) & 0.496\color{blue}(0.029) \\
        \midrule
        \multicolumn{7}{c}{unlimited games}\\
        \midrule
        \text{DQN} & 0.300\color{blue}(0.300) & 0.216\color{blue}(0.216) & 0.752\color{blue}(0.119) & 0.662\color{blue}(0.246) & 0.562\color{blue}(0.034) & 0.514\color{blue}(0.016) \\
        \text{DDQN} & 0.318\color{blue}(0.318) & 0.258\color{blue}(0.258) & 0.744\color{blue}(0.168) & 0.628\color{blue}(0.134) & 0.572\color{blue}(0.027) & 0.480\color{blue}(0.024) \\
        \text{Rainbow} & 0.316\color{blue}(0.330) & \textbf{0.280}\color{blue}(0.280) & 0.734\color{blue}(0.157) & \textbf{0.692}\color{blue}(0.157) & 0.566\color{blue}(0.017) & 0.514\color{blue}(0.014) \\
        \bottomrule
    \end{tabular}
    \caption{Agent performance on \textbf{fixed map} games. Accuracies in percentage are shown in black. We also investigate the sufficient information bonus used in our agent proposed in Section~\ref{section:reward_shaping}, which are shown in blue.}
    \label{tab:fixed_map_full}
\end{table*}

\begin{table*}[h!]
    \centering
    \small
    \begin{tabular}{r|c|c|c|c|c|c}
        \toprule
        & \multicolumn{2}{c|}{Location} & \multicolumn{2}{c|}{Existence} & \multicolumn{2}{c}{Attribute}\\
        Model & Train & Test & Train & Test & Train & Test \\
        \midrule
        \text{Human} & -- & 1.000 & -- & 1.000 & -- & 0.750 \\
        \text{Random} & -- & 0.034 & -- & 0.500 & -- & 0.499 \\
        \midrule
        \multicolumn{7}{c}{2 games}\\
        \midrule
        \text{DQN} & 0.990\color{blue}(0.990) & 0.148\color{blue}(0.162) & 1.000\color{blue}(0.779) & 0.638\color{blue}(0.157) & 1.000\color{blue}(0.039) & 0.534\color{blue}(0.033) \\
        \text{DDQN} & 0.978\color{blue}(0.978) & 0.146\color{blue}(0.152) & 1.000\color{blue}(0.727) & 0.602\color{blue}(0.158) & 1.000\color{blue}(0.043) & \textbf{0.544}\color{blue}(0.032) \\
        \text{Rainbow} & 0.916\color{blue}(0.916) & 0.178\color{blue}(0.178) & 0.972\color{blue}(0.314) & 0.602\color{blue}(0.136) & 1.000\color{blue}(0.025) & 0.512\color{blue}(0.021) \\
        \midrule
        \multicolumn{7}{c}{10 games}\\
        \midrule
        \text{DQN} & 0.818\color{blue}(0.818) & 0.156\color{blue}(0.160) & 0.898\color{blue}(0.607) & 0.566\color{blue}(0.142) & 1.000\color{blue}(0.056) & 0.518\color{blue}(0.036) \\
        \text{DDQN} & 0.794\color{blue}(0.794) & 0.142\color{blue}(0.154) & 0.868\color{blue}(0.575) & 0.606\color{blue}(0.153) & 1.000\color{blue}(0.037) & 0.500\color{blue}(0.033) \\
        \text{Rainbow} & 0.670\color{blue}(0.670) & 0.144\color{blue}(0.170) & 0.828\color{blue}(0.468) & 0.586\color{blue}(0.128) & 1.000\color{blue}(0.071) & 0.530\color{blue}(0.018) \\
        \midrule
        \multicolumn{7}{c}{100 games}\\
        \midrule
        \text{DQN} & 0.550\color{blue}(0.550) & 0.184\color{blue}(0.204) & 0.758\color{blue}(0.230) & 0.668\color{blue}(0.181) & 0.878\color{blue}(0.021) & 0.524\color{blue}(0.017) \\
        \text{DDQN} & 0.524\color{blue}(0.524) & 0.188\color{blue}(0.204) & 0.754\color{blue}(0.365) & 0.662\color{blue}(0.205) & 0.890\color{blue}(0.025) & \textbf{0.544}\color{blue}(0.019) \\
        \text{Rainbow} & 0.442\color{blue}(0.442) & 0.174\color{blue}(0.184) & 0.754\color{blue}(0.285) & 0.654\color{blue}(0.190) & 0.878\color{blue}(0.044) & 0.504\color{blue}(0.032) \\
        \midrule
        \multicolumn{7}{c}{500 games}\\
        \midrule
        \text{DQN} & 0.430\color{blue}(0.430) & 0.204\color{blue}(0.216) & 0.752\color{blue}(0.162) & 0.678\color{blue}(0.214) & 0.678\color{blue}(0.019) & 0.530\color{blue}(0.017) \\
        \text{DDQN} & 0.458\color{blue}(0.458) & 0.222\color{blue}(0.246) & 0.754\color{blue}(0.158) & 0.656\color{blue}(0.188) & 0.716\color{blue}(0.024) & 0.486\color{blue}(0.023) \\
        \text{Rainbow} & 0.370\color{blue}(0.370) & 0.172\color{blue}(0.178) & 0.748\color{blue}(0.275) & 0.678\color{blue}(0.191) & 0.636\color{blue}(0.020) & 0.494\color{blue}(0.017) \\
        \midrule
        \multicolumn{7}{c}{unlimited games}\\
        \midrule
        \text{DQN} & 0.316\color{blue}(0.316) & 0.188\color{blue}(0.188) & 0.728\color{blue}(0.213) & 0.668\color{blue}(0.218) & 0.812\color{blue}(0.055) & 0.506\color{blue}(0.018) \\
        \text{DDQN} & 0.326\color{blue}(0.326) & 0.206\color{blue}(0.206) & 0.740\color{blue}(0.246) & \textbf{0.694}\color{blue}(0.196) & 0.580\color{blue}(0.023) & 0.482\color{blue}(0.017) \\
        \text{Rainbow} & 0.340\color{blue}(0.340) & \textbf{0.258}\color{blue}(0.258) & 0.728\color{blue}(0.210) & 0.686\color{blue}(0.193) & 0.564\color{blue}(0.018) & 0.470\color{blue}(0.017) \\
        \bottomrule
    \end{tabular}
    \caption{Agent performance on \textbf{random map} games. Accuracies in percentage are shown in black. We also investigate the sufficient information bonus used in our agent proposed in Section~\ref{section:reward_shaping}, which are shown in blue.}
    \label{tab:rand_map_full}
\end{table*}

\clearpage
\begin{figure*}[t]
    \centering
    \includegraphics[width=1.0\textwidth]{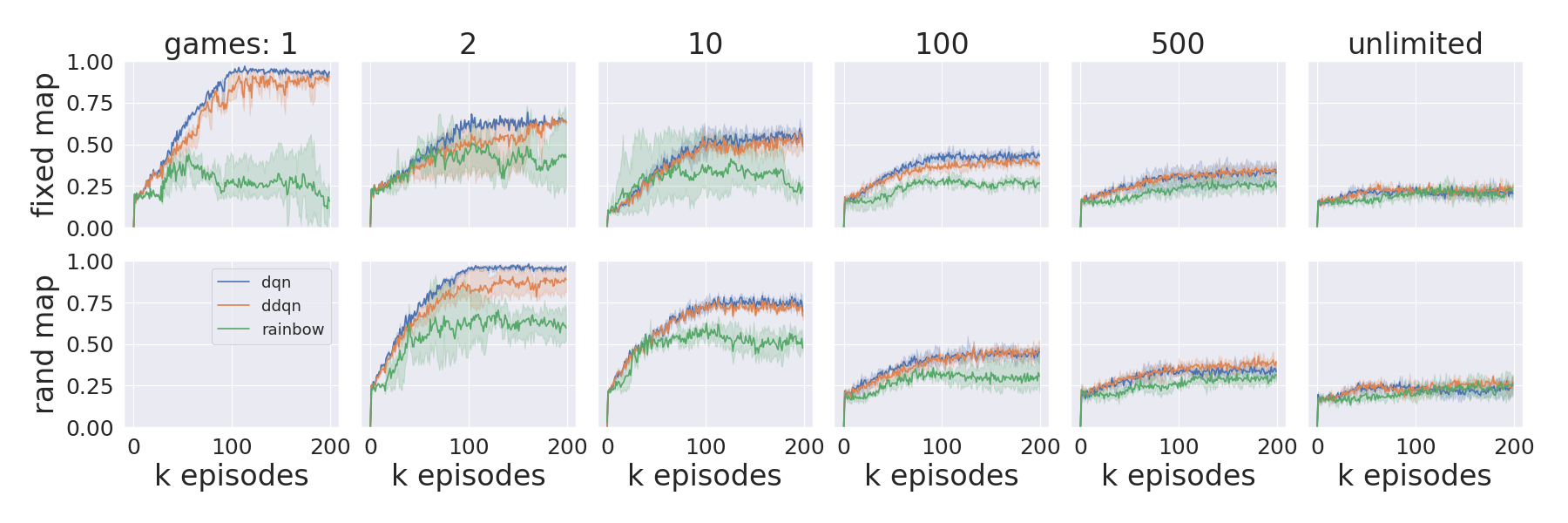}
    \caption{Training accuracy over episodes on \textbf{location} questions. Upper row: \textbf{fixed map}, 1/2/10/100/500/unlimited games; Lower row: \textbf{random map}, 2/10/100/500/unlimited games.}
    \label{fig:exp_location_full}
\end{figure*}

\begin{figure*}[t]
    \centering
    \includegraphics[width=1.0\textwidth]{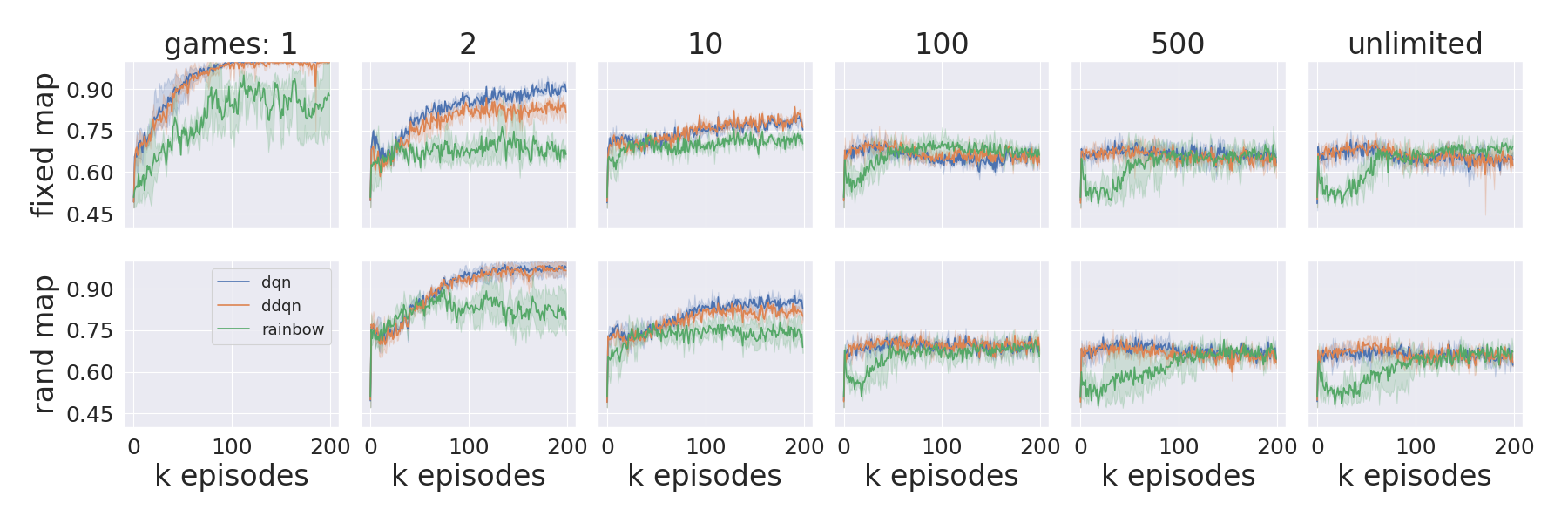}
    \caption{Training accuracy over episodes on \textbf{existence} questions. Upper row: \textbf{fixed map}, 1/2/10/100/500/unlimited games; Lower row: \textbf{random map}, 2/10/100/500/unlimited games.}
    \label{fig:exp_existence_full}
\end{figure*}

\begin{figure*}[t]
    \centering
    \includegraphics[width=1.0\textwidth]{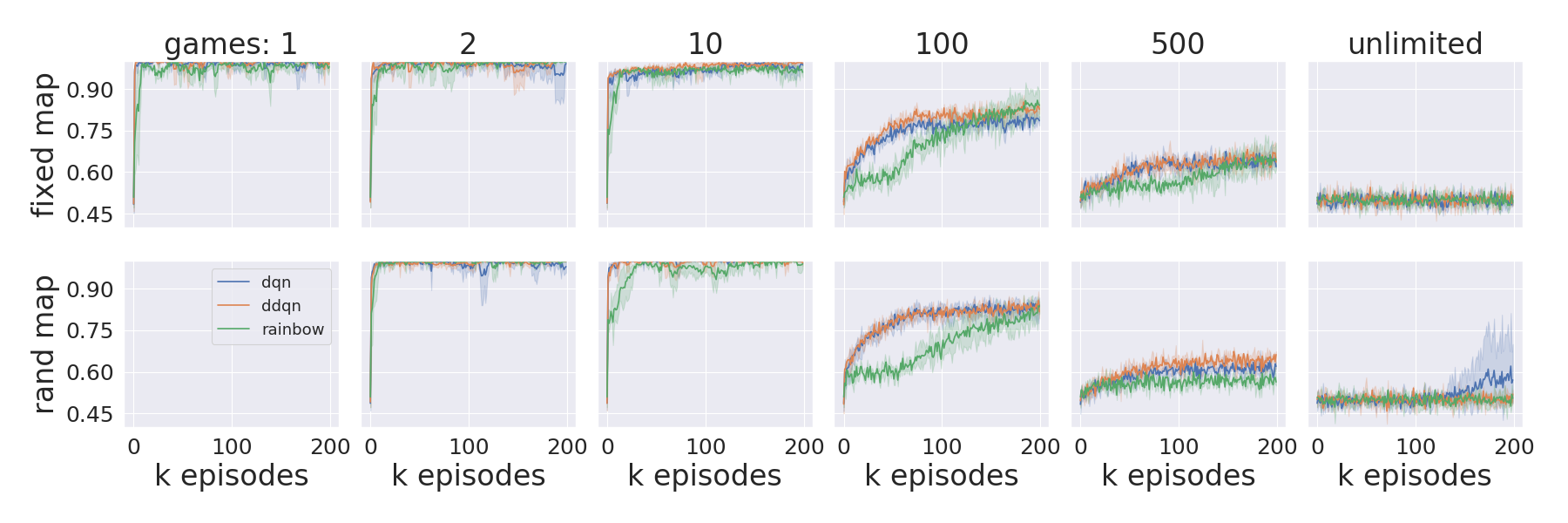}
    \caption{Training accuracy over episodes on \textbf{attribute} questions. Upper row: \textbf{fixed map}, 1/2/10/100/500/unlimited games; Lower row: \textbf{random map}, 2/10/100/500/unlimited games.}
    \label{fig:exp_attribute_full}
\end{figure*}

\end{document}